\documentclass[a4paper]{article}

\usepackage{arxiv}

\usepackage[utf8]{inputenc} 
\usepackage[T1]{fontenc}    
\usepackage{hyperref}       
\usepackage{url}            
\usepackage{booktabs}       
\usepackage{amsfonts}       
\usepackage{nicefrac}       
\usepackage{microtype}      
\usepackage{lipsum}

\usepackage{amssymb}
\usepackage{graphicx}
\usepackage{multirow}

\usepackage[capitalize,noabbrev]{cleveref}

\usepackage{float}
\floatstyle{plaintop}
\restylefloat{table}

\usepackage{gnuplot-lua-tikz}
\usepackage{subfigure}

\usepackage{caption} 
\usepackage{url}

\title{Hindi Visual Genome: A Dataset for Multimodal English-to-Hindi Machine Translation}

\author{
  Shantipriya Parida \hspace{10mm} \qquad Ond\v{r}ej Bojar\thanks{Corresponding author} \\
   Charles University, Faculty of Mathematics and Physics,\\
  Institute of Formal and Applied Linguistics,\\
 Malostransk{\'{e}} nam{\v{e}}st{\'{\i}} 25, 118 00\\ 
 Prague, Czech Republic \\
  \texttt{\{parida,bojar\}@ufal.mff.cuni.cz} \\
   \And
 Satya Ranjan Dash \\
 School of Compter Application, \\
  KIIT University, Bhubaneswar-24,\\
  Odisha, India\\
  \texttt{sdashfca@kiit.ac.in}
}

\begin{document}
\maketitle

\begin{abstract}
Visual Genome is a dataset connecting structured image information with English language.
We present ``Hindi Visual Genome'', a multimodal dataset consisting of text and
images suitable for English-Hindi multimodal machine translation task and
multimodal research. We have selected short English segments (captions) from
Visual Genome along with associated images and automatically translated them to
Hindi with manual post-editing which took the associated images into account. We
prepared a set of 31525 segments, accompanied by a challenge test set of 1400
segments. This challenge test set was created by searching for (particularly) ambiguous English words based on the embedding similarity and manually selecting those where the image helps to resolve the ambiguity.     

Our dataset is the first for multimodal English-Hindi machine translation, freely available for non-commercial research purposes. Our Hindi version of Visual Genome also allows to create Hindi image labelers or other practical tools. 

Hindi Visual Genome also serves in Workshop on Asian Translation (WAT) 2019
Multi-Modal Translation Task.
\end{abstract}


\section{Introduction}
\label{sect:introduction}
Multimodal content is gaining popularity in machine translation (MT) community due to its appealing chances to improve translation quality and its usage in commercial applications such as image caption translation for online news articles or machine translation for e-commerce product listings  \cite{multimodal:lala:et:al,ecom_multimodal:calixto:etal,imaginative:elliott:et:al,visualgrounding_mingyang:et:al}.  Although the general  performance of neural machine translation (NMT) models is very good given large amounts of parallel texts, some inputs can remain genuinely ambiguous, especially if the input context is limited. One example is the word ``mouse" in English (source) which can be translated into different forms in Hindi based on the context (e.g. either a computer mouse or a small rodent).

\begin{table}[!htb]
\caption{Hindi Visual Genome corpus details. One item consists of an English
source segment, its Hindi translation, the image and a rectangular region in the image.}\label{tab1}
\centering
\begin{tabular}{|l|r|}
\hline
Data Set &  {Items} \\
\hline
Training Set &  28,932 \\
Development Test Set (D-Test) &  998 \\
Evaluation Test Set (E-Test) & 1595 \\
Challenge Test Set (C-Test) & 1,400 \\
\hline
\end{tabular}
\label{tab:corp_details}
\end{table}

There is a limited number of multimodal datasets available and even fewer of
them are also multilingual. Our aim is to extend the set of languages available
for multimodal experiments by adding a Hindi variant of a subset of Visual Genome.

Visual Genome 
(\url{http://visualgenome.org/}, \cite{visualgenome_ranjay:et:al}) is a large set of real-world images, each
equipped with annotations of various regions in the image. The annotations
include a plain text description of the region (usually sentence parts or short sentences, e.g. ``a red ball in the air'') and also several other formally captured types of information (objects, attributes, relationships, region graphs, scene graphs, and question-answer pairs). We focus only on the textual descriptions of image regions and provide their translations into Hindi.

The main portion of our Hindi Visual Genome is intended for training purposes of
tools like multimodal translation systems or Hindi image labelers. Every item consists of an image, a rectangular region in the image, the original English caption from Visual Genome and finally our Hindi translation. Additionally, we create a challenge test set with the same structure but a different sampling that promotes the presence of ambiguous words in the English captions with respect to their meaning and thus their Hindi translation.
The final corpus statistics of the ``Hindi Visual Genome" are in \cref{tab:corp_details}. 

The paper is organized as follows: In \cref{sect:related_work}, we survey
related multimodal multilingual datasets. \cref{sec:trainset} describes the way
we selected and prepared the training set. \cref{sec:testset} is devoted to the
challenge test set: the method to find ambiguous words and the steps taken when
constructing the test set, its final statistics and a brief discussion of our
observations.
We conclude in \cref{sect:conclusion}.

Creating such a dataset enables multimodal
experimenting with Hindi for various applications
and it can also facilitate the exploration of
how the language is grounded in vision.

\section{Related Work}
\label{sect:related_work}

Multimodal neural machine translation is an emerging area where translation
takes more than text as input. It also uses features from image or sound for
generating the translated text. Combining visual features with language modeling
has shown better result for image captioning and question answering
\cite{multi_modal_context:mostafazadeh:et:al,dense_caption:yang:et:al,multimodal_caption:liu:et:al}.

Many experiments were carried out considering images to improve machine
translation, i.a. for resolving ambiguity due to different senses of words in different
contexts. One of the starting points is ``Flickr30k"
\cite{multi30k_lucia}, a multilingual (English-German, English-French, and  English-Czech) shared task based on multimodal translation was part of WMT 2018 \cite{barrault:et:al}.   
\cite{multimodal_nmt:koel:et:al} proposed a multimodal NMT system using
image feature for  Hindi-English language pair. Due to the lack of English-Hindi
multimodal data, they used a synthetic training dataset and manually curated
development and test sets for Hindi derived from the English part of Flickr30k corpus
\cite{flicker_bryan:et:al}.
\cite{wsd_with_pic:bernard:et:al} proposed a probabilistic method using pictures for word prediction constrained to a narrow set of choices, such as possible word senses. Their results suggest that images can help word sense disambiguation. 

Different techniques then followed, using various neural network architectures
for extracting and using the contextual information. One of the approaches was proposed by
\cite{multimodal:lala:et:al} for multimodal translation by replacing image
embedding with an estimated posterior probability prediction for image
categories.

\section{Training Set Preparations}
\label{sec:trainset}

To produce the main part of our corpus, we have automatically translated and
manually post-edited the English captions of ``Visual Genome'' corpus into Hindi. 

The starting point were 31525 randomly selected images from Visual Genome. Of
all the English-captioned regions available for each of the images, we randomly select
one.
To obtain the Hindi translation, we have followed these steps:

\begin{enumerate}
\item We translated all 31525 captions into Hindi using the NMT model (Tensor-to-Tensor,
\cite{t2t_ashish:et:al}) specifically trained for this purpose as
described in \cite{parida:bojar:eamt:2018}.
\item We uploaded the image, the source English caption and its Hindi machine
 translation into a  ``Translation Validation Website",\footnote{\url{http://ufallab.ms.mff.cuni.cz/~parida/index.html}} which we designed as a simple interface for post-editing the translations. One important feature was the use of a Hindi on-screen keyboard\footnote{\url{https://hinkhoj.com/api/}} 
 to enable proper text input even for users with limited operating systems.

\item Our volunteers post-edited all the Hindi translations. The volunteers were
selected based on their Hindi language proficiency.

\item We manually verified and finalized the post-edited files  to obtain the training
and test data.
\end{enumerate}

The split of the 31525 items into the training, development and test sets as
listed in \cref{tab:corp_details} was again random.

\section{Challenge Test Set Preparations}
\label{sec:testset}

\begin{figure}[t]
  \centering
  \includegraphics[scale=0.4]{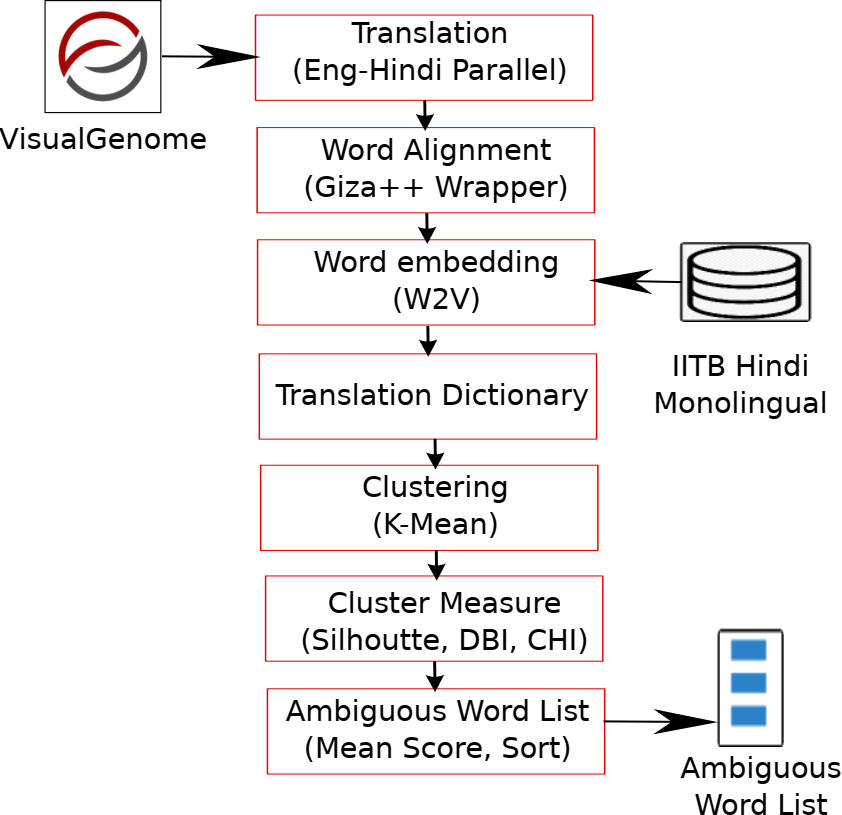}
  \caption{Overall pipeline for ambiguous word finding from input corpus.}
\label{fig:overal_experiment}  
\end{figure}

In addition to the randomly selected 31525 items described above, we prepared a
challenge test set of 1400 segments which need images for word sense
disambiguation. To achieve this targeted selection,  we first found the most ambiguous words from
the whole ``Visual Genome'' corpus and then extracted segments containing the most
ambiguous
words. The overall steps for obtaining the
ambiguous words are shown in \cref{fig:overal_experiment}. 

The detailed sequence of processing steps was as follows:

\begin{enumerate}
\item Translate all English captions from the Visual Genome dataset (3.15
millions unique strings) using a baseline machine translation systems into
Hindi, obtaining a synthetic parallel corpus. In this step, we used Google Translate.

\item Apply word alignment on the synthetic parallel corpus using GIZA++
\cite{giza_och:et:al}, in a wrapper\footnotemark{} that automatically symmetrizes two
bidirectional alignments; we used the intersection alignment.
\footnotetext{\url{https://github.com/ufal/qtleap/blob/master/cuni_train/bin/gizawrapper.pl}}

\item
\label{stepDict}
Extract all pairs of aligned words in the form of a ``translation
dictionary".
The dictionary contains key/value pairs of the English word ($E$) and all its
Hindi translations ($H_1, H_2, \dots H_n$), i.e. it has the form of the mapping
$
  E\mapsto \{H_{1},..., H_{n}\}
$.

\item \label{stepEmbs} Train Hindi word2vec (W2V) \cite{w2v_mikolov:et:al}  
word embeddings. We used the
gensim\footnote{\url{https://radimrehurek.com/gensim/tut1.html}}
\cite{rehurek:lrec:2010} implementation
and trained it on IITB
Hindi
Monolingual Corpus\footnote{\url{http://www.cfilt.iitb.ac.in/iitb_parallel/iitb_corpus_download/}}
which contains about 45 million Hindi sentences. Using such a large
collection of Hindi text improves the quality of the obtained embeddings.

\item
For each English word from the translation dictionary (see Step \ref{stepDict}), get all Hindi
translation words and their embeddings (Step \ref{stepEmbs}).

\item Apply $K$-means clustering algorithm to the embedded Hindi words to organize them according to their word similarity.

If we followed a solid definition of word senses and if we knew how many there are for a given source English word and how they match the meanings of the Hindi words, the $K$ would correspond to the number of Hindi senses that the original English word
expresses. We take the pragmatic approach and apply $K$-means for a range of values ($K$ from 2 to 6).

\item Evaluate the obtained clusters with the Silhouette Score, Davies-Bouldin Index (DBI), and Calinski-Harabaz Index (CHI)
\cite{cluster_measure_amorim:et:al,davies_bouldin:davies:et:al}. Each of the
selected scores reflects in one way or another the cleanliness of the clusters, their separation.
For the final sorting (Step \ref{stepSort}), we mix these scores using a simple average function.

The rationale behind using these scores is that if the word embeddings of the
Hindi translations can be clearly clustered into 2 or more senses, then the
meaning distinctions are big enough to indicate that the original English word
was ambiguous. The exact \emph{number} of different meanings is not too
important for our purpose.

\item
\label{stepSort}
Sort the list in descending order to get the most ambiguous words (as
 approximated by the mean of clustering measures) at the top of the list.

\item Manually check the list to validate that the selected ambiguous words
indeed potentially need an image to disambiguate them. Select a cutoff and
extract the most ambiguous English words.
\end{enumerate}

The result of this semi-automatic search and manual validation of most ambiguous
words was a list of 19 English words. For each of these words, we selected and extracted
a number of items available in the original Visual Genome and provided the same
manual validation of the Hindi translation as for the training and regular test
sets.
Incidentally, 7 images and English captions occur in both the training set and
the challenge test set.\footnote{The English segments appearing in both the
training data and the challenge test set are:
A round concert block,
Man stand in crane,
Street sign on a pole in english and chinese,
a fast moving train,
a professional tennis court,
bird characters on top of a brown cake,
players name on his shirt.
} The overlap in images (but using different regions) is larger: 359.

\cref{tab:test_data_distribution} lists the selected most ambiguous English
words and the number of items in the final challenge test set with
the given word in the English side. 
We tried to make a balance and the frequencies of the ambiguous words in the challenge test
set roughly correspond to the original frequencies in Visual Genome.

\begin{table}[t]
\centering
\begin{tabular}{||l l l||} 
 \hline
   & Word & Segment Count\\ [0.5ex] 
 \hline\hline
 1 & Stand & 180  \\ 
 2 & Court & 179 \\
 3 & Players & 137 \\
 4 & Cross & 137 \\
 5 & Second & 117 \\
 6 & Block & 116 \\
 7 & Fast & 73 \\
 8 & Date & 56 \\
 9 & Characters & 70 \\
 10 & Stamp & 60  \\
 11 & English & 42 \\
 12 & Fair & 41 \\
 13 & Fine & 45 \\
 14 & Press & 35 \\
 15 & Forms & 44 \\
 16 & Springs & 30 \\
 17 & Models & 25 \\
 18 & Forces & 9 \\
 19 & Penalty & 4 \\
\hline
    & Total & 1400 \\
 \hline
\end{tabular}
\caption{Challenge test set: distribution of the ambiguous words.}
\label{tab:test_data_distribution}
\end{table}

\begin{figure*}[!htb]
\centering
\subfigure[Street sign advising of penalty.]{
   \includegraphics[width=.45\textwidth]{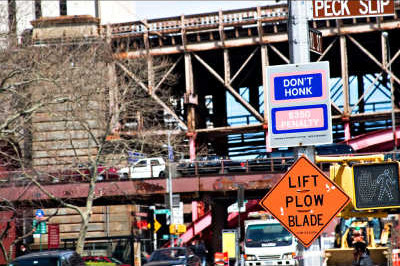}
    \label{penalty_fine}
}
\subfigure[The penalty box is white lined.]{
   \includegraphics[width=.45\textwidth]{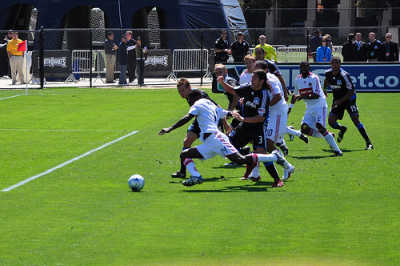}
    \label{penalty_kick}
}
\caption{An illustration of two meanings of the word ``penalty'' exemplified
with two images.}
\label{fig:manual_analysis_ambiguity}
\end{figure*}

\cref{fig:manual_analysis_ambiguity} illustrates two sample items selected for
the word ``penalty'' (Hindi translation omitted here). We see that for humans, the images
are clearly disambiguating the meaning of the word: the fine to be paid for
honking vs. the kick in a soccer match.

Arguably, the surrounding English words
in the source segments (e.g. ``street'' vs. ``white lined'') can be used by
machine translation systems to pick the correct translation even without access
to the image. The size of the original dataset of images with captions however
did not allow us to further limit the selection to segments where the text alone
is not sufficient for the disambiguation.

\section{Conclusion and Future Work}
\label{sect:conclusion}

We presented a multimodal English-to-Hindi dataset. To
the best of our knowledge, this is the first such dataset that includes an
Indian language. The dataset can serve e.g.
in Hindi image captioning but our primary intended use case was
research into the employment of images as additional input to improve machine translation
quality.

To this end, we created also a dedicated challenge test set with text segments
containing ambiguous words where the image can help with the disambiguation.
With this goal, the dataset also serves in WAT 2019\footnote{\url{http://lotus.kuee.kyoto-u.ac.jp/WAT/WAT2019/index.html}}
shared
task on multi-modal translation.\footnote{\url{https://ufal.mff.cuni.cz/hindi-visual-genome/wat-2019-multimodal-task}}

We
illustrated that the text-only information in the surrounding words could be sufficient for the
disambiguation. One interesting research direction would be thus to ignore all
the surrounding words and simply ask: given the image, what is the correct Hindi
translation of this ambiguous English word. Another option we would like to
pursue is to search larger datasets for cases where even the whole segment does
not give a clear indication of the meaning of an ambiguous word.

Our ``Hindi Visual Genome" is available for research and non-commercial use under a Creative Commons Attribution-NonCommercial-ShareAlike 4.0 License\footnote{\url{https://creativecommons.org/licenses/by-nc-sa/4.0/}} at \url{http://hdl.handle.net/11234/1-2997}.


\section{Acknowledgments}

We are grateful to Vighnesh Chenthil Kumar, a summer intern from IIIT Hyderabad
at Charles University for his help with the semi-automatic search for the most
ambiguous words. The work was carried out during Shantipriya Parida's post-doc
funded by Charles University.

This work has been supported by
the grants 19-26934X (NEUREM3) of the Czech Science Foundation 
and ``Progress'' Q18+Q48 of Charles University, and using language resources distributed by the LINDAT/CLARIN project of the Ministry of Education, Youth and Sports of the Czech Republic (projects LM2015\discretionary{-}{}{}071 and OP VVV VI CZ.02.1.01/0.0/0.0/16013/0001781).

\bibliographystyle{unsrt}

\end{document}